\documentclass{llncs}  

\usepackage{graphicx} 
\usepackage[T1]{fontenc}
\usepackage[english]{babel}
\usepackage{array}
\usepackage{tabularx}
\usepackage{booktabs} 
\usepackage{makeidx}  % allows for index generation
\usepackage{xparse}
\usepackage{moredefs}
\usepackage{lips}
\usepackage{epstopdf} 
\usepackage{csquotes}
\usepackage{xcolor}
\usepackage{times}
\usepackage[hidelinks]{hyperref}
\usepackage[nolist,nohyperlinks]{acronym}
\usepackage{comment}
\usepackage{paralist}
\usepackage{wrapfig}
\usepackage{multirow}
\usepackage{microtype}
\usepackage{listings}
\usepackage[super]{nth}
\usepackage{natbib}
\usepackage{esvect}
\usepackage{amsmath}
\usepackage{cleveref}
\usepackage{makecell}
\usepackage{siunitx}

\usepackage{float}
\floatstyle{plaintop}
\restylefloat{table}

\usepackage[style=base,labelfont=bf,labelsep=period,tableposition=top,font=small]{caption}
\makeatletter
\renewcommand\@biblabel[1]{#1.}
\makeatother
\addto\captionsenglish{}

\usepackage{fancyhdr}
\fancypagestyle{WI_footer}{\fancyhf{}\fancyfoot[L]{\color{black}\scriptsize 20th International Conference on Wirtschaftsinformatik\\September 2025, Münster, Germany}}

\crefformat{footnote}{#2\footnotemark[#1]#3}
\expandafter\def\expandafter\UrlBreaks\expandafter{\UrlBreaks%  save the current one
  \do\a\do\b\do\c\do\d\do\e\do\f\do\g\do\h\do\i\do\j%
  \do\k\do\l\do\m\do\n\do\o\do\p\do\q\do\r\do\s\do\t%
  \do\u\do\v\do\w\do\x\do\y\do\z\do\A\do\B\do\C\do\D%
  \do\E\do\F\do\G\do\H\do\I\do\J\do\K\do\L\do\M\do\N%
  \do\O\do\P\do\Q\do\R\do\S\do\T\do\U\do\V\do\W\do\X%
  \do\Y\do\Z}

\newcolumntype{L}[1]{>{\raggedright\arraybackslash}p{#1}}   % linksbündig mit Breitenangabe
\newcolumntype{C}[1]{>{\centering\arraybackslash}p{#1}}     % zentriert mit Breitenangabe
\newcolumntype{R}[1]{>{\raggedleft\arraybackslash}p{#1}}    % rechtsbündig mit Breitenangabe

\begin{document}
\frontmatter          % for the preliminaries

\mainmatter

\title{Unveiling Location-Specific Price Drivers: A Two-Stage Cluster Analysis for Interpretable House Price Predictions}

\subtitle{Research Paper} %Specify type of research paper here!

\author{Paul Gümmer\inst{1,2} \and
Julian Rosenberger\inst{3} \and
Mathias Kraus\inst{3} \and
Patrick Zschech\inst{2,4} \and
Nico Hambauer\inst{3}
}

\institute{University of Vienna, Vienna, Austria \\ \email{paulg46@univie.ac.at} \and Leipzig University, Leipzig, Germany \\
\and University of Regensburg, Regensburg, Germany \\ \email{\{julian.rosenberger, mathias.kraus, nico.hambauer\}@ur.de} 
\and TU Dresden, Dresden, Germany \\ \email{patrick.zschech@tu-dresden.de}
}

% -----------------------
% |  Begin of Document  |
% -----------------------
\maketitle
\setcounter{footnote}{0}

% ------------- 
% |  Abstract and Keywords  |
% -------------
\begin{abstract}
%Überarbeiteter Vorschlag für den Abstract (150 words):
House price valuation remains challenging due to localized market variations. Existing approaches often rely on black-box machine learning models, which lack interpretability, or simplistic methods like linear regression (LR), which fail to capture market heterogeneity. To address this, we propose a  machine learning approach that applies two-stage clustering, first grouping properties based on minimal location-based features before incorporating additional features. Each cluster is then modeled using either LR or a generalized additive model (GAM), balancing predictive performance with interpretability. Constructing and evaluating our models on \num{43\,309} German house property listings from 2023, we achieve a $36\,\%$ improvement for the GAM and $58\,\%$ for LR in mean absolute error compared to models without clustering. Additionally, graphical analyses unveil pattern shifts between clusters. These findings emphasize the importance of cluster-specific insights, enhancing interpretability and offering practical value for buyers, sellers, and real estate analysts seeking more reliable property valuations.
\end{abstract}


\thispagestyle{WI_footer}

% \noindent Please note that the review process is double-blind. Manuscripts submitted for review MUST NOT include author information---neither on the title page nor in the page header, etc. 
% This paragraph is intended to serve as filler in your initial submission to keep the overall length of the paper consistent, even if the author information section above is not included. You can delete it later in the final submission when you add the actual author information.

% ------------- 
% |  Content  |
% -------------

\section{Introduction}
\label{sec:introduction}

The German housing market faces unprecedented challenges in price prediction. House price predictions remain challenging due to numerous factors including market heterogeneity, location characteristics, and diverging feature effects. 
While machine learning promises more accurate valuations, many current approaches either rely on simple linear regression (LR) or on overly complex \textit{black-box} algorithms that offer little interpretability, making their decision logic opaque \citep{lorenz2023interpretable, janiesch_machine_2021}. Modeling heterogeneous markets with either overly complex models or with simple linear models presents a dilemma. Practitioners must choose between highly accurate, but less explainable predictions from complex models, or more transparent but less accurate estimates from simple linear models \citep{hong2020house, kruschel2025challenging}. 

Traditionally, machine learning approaches apply models across an entire dataset, treating it as a single problem, respectively assuming homogeneous feature effects across samples \citep{hong2020house}. We therefore refer to them as global machine learning models. On the contrary, house price prediction challenges arise from heterogeneous data representing diverse local markets. Therefore, developing methodologies that accommodate this market granularity is essential for robust predictive modeling.

Recent advances in machine learning demonstrate that clustering techniques can improve prediction performance by accounting for market heterogeneity and individuality \citep{azimlu2021house, hambauer_benchmarking_2025}. These approaches overcome the limitations of global models that employ one-size-fits-all methodologies. However, existing house pricing research typically implements clustering merely as a preprocessing step, subsequently compromising interpretability by employing complex models in a secondary phase \citep{mayer2019estimation}. What remains unexplored is an integrated approach that leverages clustering to enhance both prediction quality and model interpretability simultaneously, thus resolving the apparent trade-off between transparent and accurate models.
To address this gap, we formulate the following research question:

\medskip
\noindent\textbf{\textit{Research Question:}}
\emph{``Can a clustering approach improve the predictive performance of interpretable models for house price prediction?''}
\medskip 

Our paper addresses this question by proposing a fine-granular interpretable cluster analysis that decomposes the complex pricing problem into smaller, manageable subproblems. Our two-stage clustering approach first groups properties by key location features, then applies refined clustering with additional property characteristics. For each cluster, we train LR and Explainable Boosting Machine (EBM) models. EBM is a generalized additive model (GAM) from the InterpretML package \citep{nori2019interpretml}, which maintains interpretability while capturing non-linear feature effects through intuitive shape plots \citep{kruschel2025challenging,lou2013accurate}. Unlike linear approaches, EBM reveals complex data patterns, enabling detailed feature effect visualization as demonstrated in \Cref{subsec:assessment_of-price_effects}. We evaluate our approach on \num{43\,309} German house property listings from 2023. The main contributions of this paper are:
\begin{enumerate}
    \item A two-stage clustering approach that uses location-based and property-based features in each clustering stage respectively.
    \item A comparative performance assessment of two interpretable models at cluster-level: LR and EBM.
    \item Empirical evidence showing substantial performance improvements of $58\,\%$ for LR and $36\,\%$ for EBM when using the proposed two-stage clustering approach compared to global models.
    \item A visual assessment and interpretation of model outputs using the EBM, which reveals cluster-specific price effects across different location clusters.
\end{enumerate}

Our findings have important implications for both practice and research: practitioners gain more reliable and interpretable valuations, while researchers benefit from a new methodological approach that bridges the gap between prediction performance and interpretability in real estate valuation. The rest of this paper is structured as follows: \Cref{sec:related_work} outlines related work that use clustered and locally weighted approaches. \Cref{sec:experimental-setup} reports our experimental setup, whereas \Cref{sec:method} explains our two-staged clustered analysis. \Cref{sec:results} reports the results and \Cref{sec:discussion} discusses our findings and outlines implications for researchers and practitioners. Finally, \Cref{sec:conclusion} summarizes our paper with concluding remarks.

\section{Related Work}
\label{sec:related_work}    

Recent years have seen a shift towards machine learning in real estate valuation \citep{park2015using, ho2021predicting}. Traditional valuation methods like the comparison approach, income approach, and cost approach are often limited, especially when considering complex relationships and spatial effects in housing markets. Additionally, these methods struggle to capture non-linear feature effects and interactions between features. Such types of models, which estimate property values based on a few constituent features, are referred to as hedonic pricing models. They often suffer from functional form misspecification when the assumed relationship between features and price does not capture the true underlying dynamics \citep{root2023review}.

Earlier studies in house price predictions demonstrated that neural networks achieved higher performance than traditional regression models \citep{din2001environmental, peterson2009neural}. However, they also tend to require a larger amount of samples in the dataset and imply a more intricate model parametrization. While neural networks effectively addressed form misspecifications of hedonic pricing models, they also introduced challenges, such as overfitting and limited interpretability in identifying key price determinants \citep{hui2009property}. As an alternative, geographically weighted regression (GWR) was introduced to enhance traditional regression models by allowing location-specific coefficients. GWR proved particularly valuable for mass property valuation by balancing performance with interpretability \citep{mccluskey2012potential}. Further research compared GWR with GAMs and ordinary least squares (OLS) regression using German rental listings from 2013 to 2015 \citep{cajias2018spatial}. Their findings indicated that simpler OLS models yielded superior results, as GWR weights remained constant over time, potentially resulting in suboptimal predictions in dynamic markets, which highlights the importance of model simplicity.

Further developments showed that random forest (RF) models outperformed LR in predicting apartment prices in South Korea \citep{hong2020house}. Similarly, an evaluation of six different valuation methods using over \num{123\,000} single-family home transactions in Switzerland identified Gradient Boosting as the most effective approach \citep{mayer2019estimation}. These studies highlighted the importance of regularization in tree-based models to avoid prediction bias and overfitting in real estate price predictions \citep{mullainathan2017machine}. More recently, deep learning models have gained traction in property price estimation. A comparative study on deep learning applications demonstrated that transformer models and RF outperformed traditional regression-based approaches in predictive performance, while long short-term memory networks and gated recurrent units performed poorly, suggesting that not all deep learning methods are well-suited for real estate price forecasting \citep{shi2023deep}. This research highlights the potential of feature selection and dimensionality reduction in improving model performance, emphasizing the necessity of balancing complexity with interpretability.

A significant advancement came with the introduction of market segmentation before applying regression models \citep{azimlu2021house}. By first clustering properties using $k$-means and then applying regression models to each group, this approach achieved improved predictive performance, especially for datasets with high variance. This suggested that different market segments might require different prediction models.

Recent research has also focused on identifying key price-influencing factors. Processing location data plays a crucial role \citep{alfaro2020fully} and environment factors such as proximity to schools, public transportation, and shopping facilities are highly predictive \citep{wang2021deep}. Additionally interpretable machine learning methods demonstrated that property age and size were major factors in pricing, with feature combinations showing more impact than individual characteristics \citep{lorenz2023interpretable}.

Next to traditional real estate features, location data, and environmental factors, alternative data sources have been explored to enhance real estate price prediction. \cite{sun2014combining} proposed an integrated approach combining online news articles and web search behavior to forecast real estate price fluctuations. The authors find that incorporating search engine query data alongside sentiment analysis from news articles improved predictive performance, suggesting that consumer behavior and market sentiment are important factors in real estate valuation.

Despite these advancements, a gap remains between interpretable but less accurate linear models and complex but opaque machine learning approaches like RF. While the approach of clustering properties before applying interpretable machine learning models shows promise, this combination remains underexplored \citep{hong2020house}. It is still unclear which features substantially influence price in each cluster and whether they differ across market segments. Combining clustering with interpretable models could lead to not only more accurate predictions but also more actionable insights into specific features affecting property prices in various market segments.

\section{Experimental Setup}
\label{sec:experimental-setup}

In our experiments, data is gathered, examined, and preprocessed. After that, the data is grouped into clusters. During the data analysis phase, different clustering methods and algorithms are employed to categorize house property listings into groups that are more uniform, based on price and features. Then, the outcomes of various methods and clusters are compared to identify the characteristics of each cluster. 

Subsequently, the cluster-level models undergo a training phase respectively. We employ both a LR model and EBM interchangeably. This allows us to compare these interpretable approaches in our experimental results. Since the EBM is a GAM and therefore captures feature-wise non-linear effects, it is presumed to excel in identifying more nuanced patterns for each cluster, thus enabling us to examine in more depth how these features drive the house price. Based on performance measures we comparatively assess whether and to what extent the clustering methods enhance the predictive performance of real estate price predictions. Moreover, we combine cluster-level models in shared visualizations to compare the patterns between location clusters using the EBM.

\subsection{Data Acquisition}

This study uses a comprehensive dataset from the RWI-Leibniz Institute for Economic Research, entitled RED Campus dataset in its fifth version, provided by ImmobilienScout24 for non-commercial use. The dataset includes location information mainly in the form of postal codes of the properties, detailed property characteristics, and price data. The dataset has been acquired, validated, with some preprocessing steps by the Research Data Center (FDZ) Ruhr \citep{schaffner2024fdz}.

The dataset in its fifth version is divided into two separate files: the Cross-Section dataset, which contains real estate listings from 2023 across Germany, and the Panel Campus dataset, which includes listings from the 15 largest cities over time between 2008-2023. The Cross-Section dataset differentiates between listings for house sales, apartment sales, and apartment rentals. We focus on house sale listings from the Cross-Section dataset, thereby analyzing a substantial part of the recent german house market.

\subsection{Data Preprocessing}
 
First, data cleaning was performed to avoid overfitting and skewing the analysis. This included removing samples that did not have the \textit{price} target and removing duplicate samples that were likely re-published on the platform. To account for slight variations in price and area when re-published, we removed duplicates differing by up to $5\,\%$ in these features. We removed properties with a plot size less than $30$ square meters. Moreover, samples with unrealistically high feature values were also removed manually. Further than that, no winsorization or outlier removal was necessary nor performed. The dataset resulting from acquisition and preprocessing had \num{43\,309} entries.

Second, to improve the representation of geographical information, latitude and longitude features were derived from the postal codes. While postal codes provide location data, they are not always geographically sequential. By converting postal codes into latitude and longitude coordinates, we obtained continuous numerical features that more accurately reflect the properties' actual locations and distances between them.

Third, irrelevant features and those with more than $90\,\%$ missing values were removed \citep{schaffner2024fdz}. Missing values in key features were handled as follows: 1)~\textit{monthly rental income} values were set to zero, assuming the property was not rented in case no income was reported. 2)~for the feature \textit{last renovation year}, missing values were replaced by the property's construction year, indicating no renovations were performed since construction. 3)~missing counts of bathrooms and bedrooms were estimated using LR based on the total number of rooms, reflecting their proportional relationship.

Fourth, feature selection was performed to reduce dimensionality and enhance model performance: An initial selection removed features irrelevant to price prediction, such as metadata about the listing (e.g., number of views). Features with high missing values that could not be imputed were excluded \citep{schaffner2024fdz}. The number of location features was reduced, retaining postal code, latitude and longitude as the primary identifiers. The EBM model was leveraged to identify and retain the most important features influencing the property price \citep{lou2013accurate}. We resulted in a final set of 18 features. A list of these features will be made publicly in our online appendix.\footnote{\url{https://osf.io/jh2kb/?view}}

Lastly, categorical features were encoded to be suitable for our models. Binary features were encoded as 0 (No) and 1 (Yes). Ordinal features with a natural order (e.g., energy efficiency class, property condition) were label-encoded with integers reflecting their order \citep{muller2016introduction}. Nominal features without a natural order (e.g., heating type, property type) were one-hot encoded to avoid introducing ordinal relationships where none exist \citep{muller2016introduction}.

\subsection{Evaluation and Validation}

Model performance and clustering effectiveness were assessed using mean absolute error (MAE) and root mean square error (RMSE). MAE measures the average magnitude of errors in predictions, providing an intuitive understanding of predictive performance. RMSE highlights larger errors due to the squaring of residuals, offering insight into the models prediction outliers \citep{chai2014root}. K-fold cross-validation was employed to ensure the robustness of the models and prevent biases due to data splitting. Feature effects were visualized using the EBM model, as it reveals feature-wise non-linear feature effects that enable us to derive actionable insights. This enhances the results section by complementing model performance with model explanations in \Cref{subsec:assessment_of-price_effects}.

\section{Method}
\label{sec:method}

The novelty of our method lies in the application of a granular two-stage clustering approach, followed by the use of interpretable models at the cluster-level. The first clustering stage uses a small feature set, while the second clustering stage uses the entire feature set. In the first stage, mainly latitude, and longitude are used to group geographically proximate houses with similar price levels combined with relying on information from the \textit{price} target. In the second stage, a more nuanced clustering approach with a broader feature set is employed. In the final prediction stage, our method proposes to make use of simple models, namely LR and EBM, to make house price predictions more interpretable. We make our implementation freely available for reproduction and application in other contexts.\footnote{\url{https://gitlab.com/house-price-predicition/cluster-analysis-for-interpretable-house-price-predictions}}

% neu von hier
For clustering, we use regression trees, $k$-means, and $k$-nearest neighbors (KNN), which are among the most frequently applied clustering approaches \citep{hambauer_benchmarking_2025}.
Regression trees split the feature space via thresholding and assign samples to their cluster via leaf-modeling \citep{de2018new}. $K$-means and KNN identify $k$ centroids, by searching for dense data regions in the feature space \citep{friedman1975algorithm}.

For the first stage we chose $k$-means as it revealed the best results for $k=2$. For KNN and regression trees we would only indirectly set a maximum cluster size through other hyperparameters. Therefore, KNN and regression trees produced a larger number of clusters, which made us disregard them for the first stage. We deliberately opted for only two clusters in this initial clustering stage, as each additional cluster would have increased the number of sub-clusters by a factor of eight, which would reduce interpretability and human comprehensibility.

In the second stage, we compared $k$-means with a regression tree.
Guided by the regression tree, we fixed the tree depth at three, which yields eight terminal leaves. A depth of two produced clusters that were almost indistinguishable, whereas a depth of four generated 16 leaves, many of them contained too few observations to build stable cluster-specific models. To ensure a fair comparison, we therefore also set $k=8$ for the $k$-means variant. We used the Elbow method to validate that, for the set of features, 8 clusters are feasible (see online appendix).\footnote{\url{https://osf.io/jh2kb/?view}} The plot shows diminishing returns starting around 6-8 clusters, with 8 clusters providing a good balance between model complexity and within-cluster variance reduction.
Our guiding principle throughout was to limit the number of clusters to a scale that remains interpretable for human analysts while still capturing the key heterogeneity in the data.
% neu bis hier

Based on their interpretability, two models were selected: A simple LR using an L1 penalty and a slightly more complex yet interpretable non-linear GAM, specifically the EBM \citep{lou2013accurate}. LR performs feature selection by penalizing the absolute size of the coefficients using an L1 penalty, effectively reducing less important feature weights to zero and preventing overfitting \citep{tibshirani1996regression}. The EBM is an advanced non-linear model that combines predictive power with inherently interpretability to capture complex relationships without becoming a \textit{black-box} and demonstrated superior performance over traditional models for various problems \citep{kruschel2025challenging}.

\section{Results}
\label{sec:results}

In this section, we systematically present our results. First, we compare the predictive performance of a global model without clustering against three different clustering approaches. These three approaches include a simple approach using location-based $k$-means clustering, and two variations of repeated two-stage clustering with variants of both $k$-means and regression trees. 

\subsection{Assessment of Predictive Performance}
\label{subsec:predictive-performance}

\Cref{tab:model_clustering_comparison} illustrates the comparative assessment of predictive performance for both LR and the EBM. All performance measures, respectively errors are in Euros (€). In all four approaches, the EBM achieves better performance than LR. Nevertheless, LR as part of various clustered approaches shows similar performance compared to the EBM while still being intrinsically easier to interpret, due to the small set of coefficients. Moreover, both prediction models perform substantially better once the data is clustered.

\begin{table}[h!]
\caption{Assessment of predictive performance for clustering approaches under mean absolute error (MAE) and root mean squared error (RMSE) in Euros. Best approach is underlined.}
\label{tab:model_clustering_comparison}
\centering
\renewcommand{\arraystretch}{1.2} % Adjusts row height
\setlength{\tabcolsep}{6pt} % Adjusts column spacing
\resizebox{1.0\linewidth}{!}{
\begin{tabular}{lp{0.3\linewidth}llrr}
\toprule
Model & Approach & \multicolumn{2}{l}{Clustering Phases} & \makecell[r]{Average MAE \\(in €)} & \makecell[r]{Average RMSE \\(in €)} \\ \cmidrule(lr){3-4}
& & 1\textsuperscript{st} Phase & 2\textsuperscript{nd} Phase & & \\ \midrule
EBM & No Clustering & - & - & \num{110\,387} & \num{165\,509} \\ \cmidrule(lr){2-6}
    & Location clustering & $k$-means & - & \num{87\,276} & \num{115\,627} \\
    & Two-stage clustering & $k$-means & regression tree & \num{88\,696} & \num{115\,291} \\
    & Two-stage clustering & $k$-means & $k$-means & \underline{\num{80\,925}} & \underline{\num{104\,189}} \\
\midrule
\makecell[l]{LR} & No Clustering & - & - & \num{170\,745} & \num {244\,552} \\ \cmidrule(lr){2-6}
    & Location Clustering & $k$-means & - & \num{127\,272} & \num{160\,076} \\
    & Two-stage clustering & $k$-means & regression tree & \num{122\,295} & \num{151\,510} \\
    & Two-stage clustering & $k$-means & $k$-means & \num{107\,996} & \num{134\,297} \\
\bottomrule
\end{tabular}
}
\end{table}

Location-based clustering brings the largest relative performance difference looking at LR without clustering (\num{170\,745} MAE) vs. location clustering (\num{127\,272} MAE) and EBM without clustering ($110\,387$ MAE) vs. location clustering (\num{87\,276} MAE). For the second stage of clustering which goes beyond just location, by using various features for clustering, applying $k$-means twice in both clustering stages is the most effective for both models. This further improved the prediction performance comparing LR with only location clustering (\num{127\,272} MAE) vs. clustering twice ($104\,189$ MAE) and EBM with only location clustering (\num{87\,276} MAE) vs. clustering twice (\num{80\,925} MAE).
Overall the two-stage approach improves predictive performance by approximately $36\%$, looking at EBM without clustering (\num{110\,387} MAE) and two-stage clustering with $k$-means (num{80\,925} MAE) as well as approximately $58\%$ looking at LR without clustering (\num{170\,745} MAE) vs. two-stage clustering with $k$-means (\num{107\,996} MAE).

\subsection{Visualizations of Location-based Assessment}

\begin{figure}[h]
    \centering
    \includegraphics[width=1\linewidth]{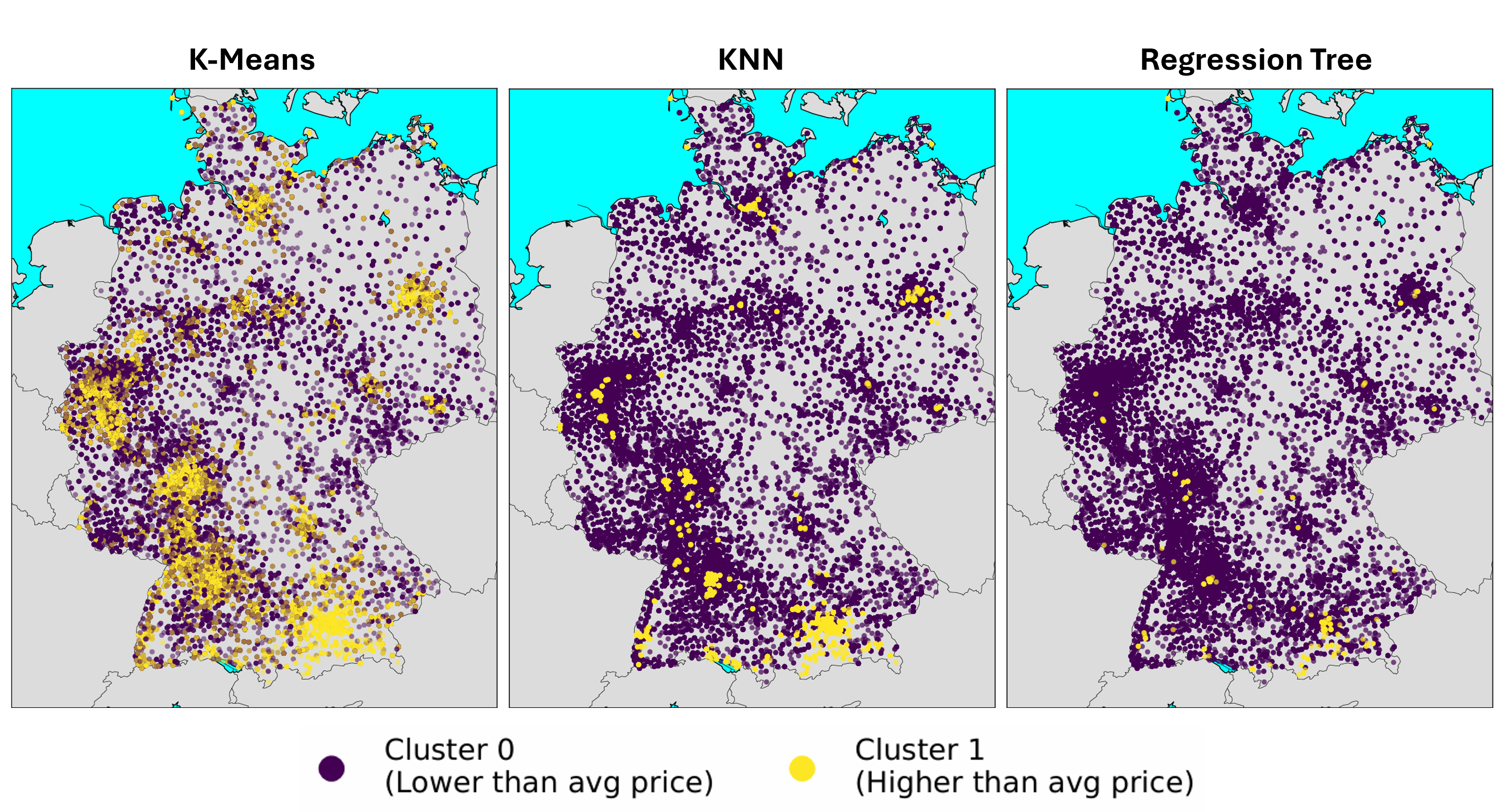}
    \caption{Results of the location clusters for different algorithms}
    \label{fig:location_based_clustering}
\end{figure}

Location-based clustering proved to have a large impact on the predictive performance. In the previous analysis we relied on $k$-means, while alternative clustering approaches would also be possible. Therefore, we applied the location-based clustering approach using three different algorithms on a limited set of features and visualize the results in \Cref{fig:location_based_clustering} on a map of Germany.

The left part of \Cref{fig:location_based_clustering} visualizes the clustering result obtained by using $k$-means. A $K$-Nearest-Neighbor (KNN) regressor is used to predict property prices based on the location, which are then grouped into two clusters using binning (middle part of \Cref{fig:location_based_clustering}). The right part of \Cref{fig:location_based_clustering} visualizes the clusters provided by a pruned regression tree with a maximum depth of fifteen aggregated by setting a threshold for the leaf nodes. The differences between the outcome of this analysis become especially obvious with the $k$-means algorithm, which distinctly identifies major metropolitan areas such as Hamburg, Berlin, Munich, and Frankfurt am Main. Based on the graphical representation, $k$-means provides the best separation between more expensive and less expensive properties when only two clusters are used.

\subsection{Assessment of Cluster-specific Price Effects}
\label{subsec:assessment_of-price_effects}

The application of location-based clustering effectively reduced the dominant influence of location in the subsequent sub-clustering step, allowing other property characteristics to become more prominent in the modeling phase using LR and EBM. An analysis of the feature relationships learned by the EBM model reveals that the same feature can have different effects on price depending on the respective location cluster and sub-clusters, underscoring the heterogeneity of market segments.

\begin{figure}[h]
    \centering
    \includegraphics[width=0.9\linewidth]{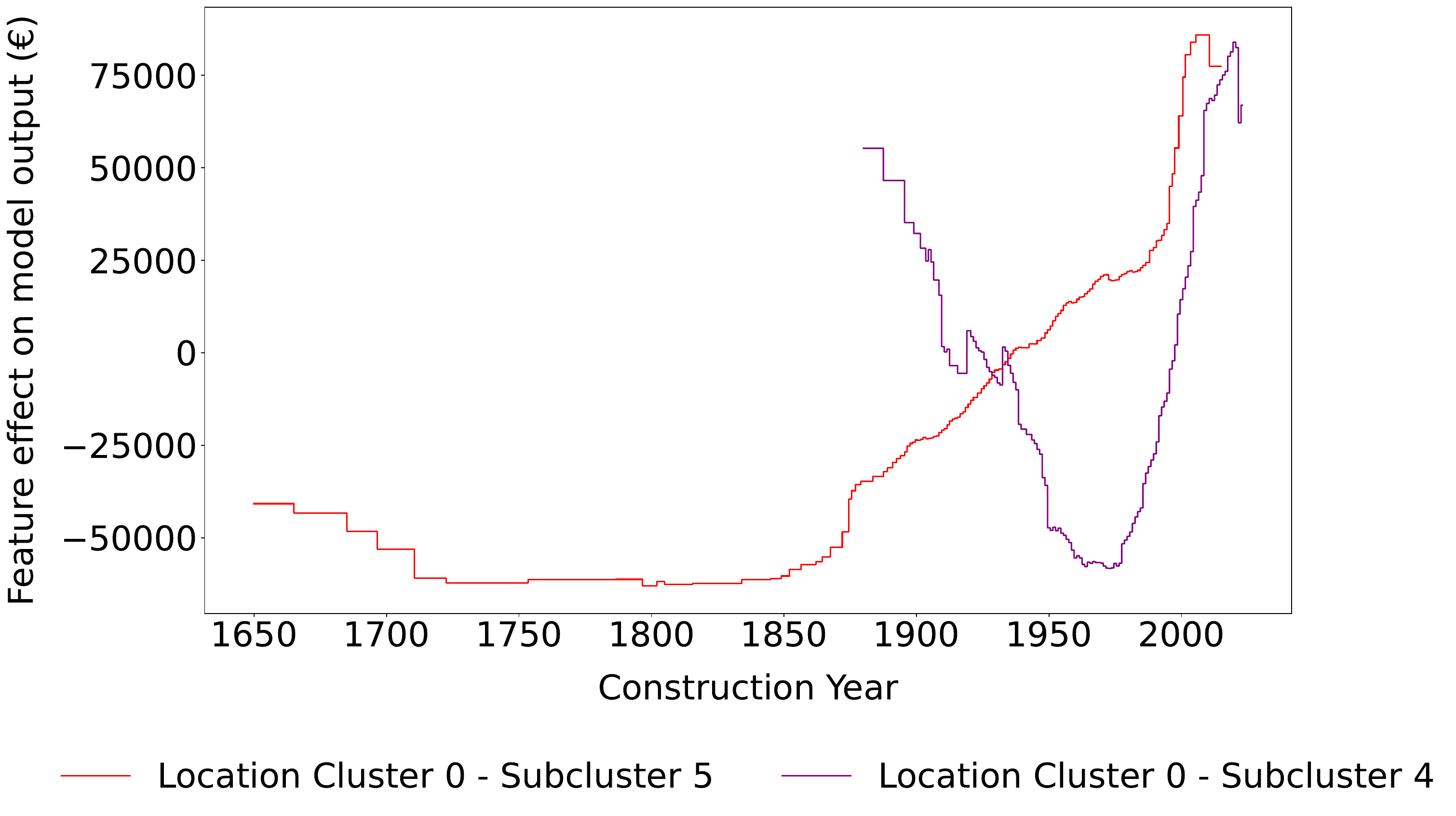}
    \caption{Feature effect of construction year on the price using the EBM model based on two distinct clusters respectively.}
    \label{fig:_combined_ebm-construction_year-price}
\end{figure}

\Cref{fig:_combined_ebm-construction_year-price} illustrates the influence of \textit{construction year} on property prices in two distinct sub-clusters. Notably, the different sub-clusters have distinct feature ranges due to the clustering. The line in purple representing sub-cluster 4 begins at a later \textit{construction year} since that cluster only includes samples that have the feature \textit{construction year} around that feature range. The cluster-specific EBM model learns a different relationship for properties in sub-cluster 4 compared to sub-cluster 5. For instance, at \textit{construction year} 1950, there's approximately a \num{70\,000}€ difference in feature effect between the two sub-clusters, with sub-cluster 4 showing a negative effect on price while sub-cluster 5 demonstrates a positive effect in that area. Moreover, comparing sub-cluster 5 to sub-cluster 4 in general, we can see that in sub-cluster 5 older houses have a high negative influence of approximately \num{62\,500} between 1720 until approximately 1850, after which the influence increases continuously up to the year 2000, reaching a positive feature effect by roughly 1930. In sub-cluster 4 however, houses built around 1900 start with having a positive influence, which decreases until about 1975 and then rises again. These patterns suggest a specific preference for historical, more luxurious properties in sub-cluster 4, while sub-cluster 5 could include more standardized houses. 

\begin{figure}[ht]
    \centering
    \includegraphics[width=0.9\linewidth]{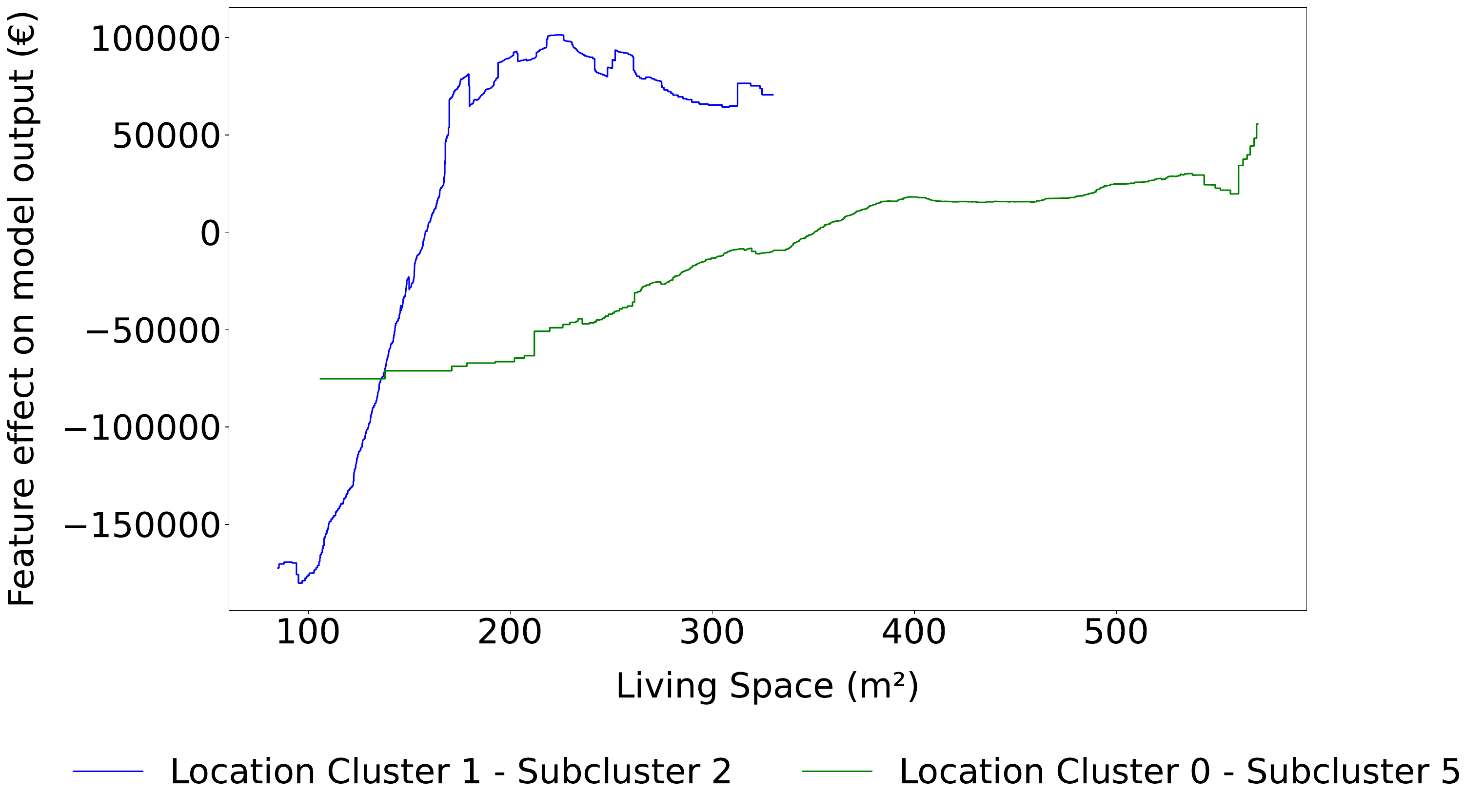}
    \caption{Feature effect of living space on the price using the EBM model based on two distinct clusters respectively.}
    \label{fig:_combined_ebm-living_space-price}
\end{figure}

\Cref{fig:_combined_ebm-living_space-price} depicts the effect of \textit{living space} on house property prices. In sub-cluster 2 prices rise with increasing square meters up to a peak at approximately $175$ square meters, remaining roughly constant afterwards. In contrast, in sub-cluster 5 the price increases slowly, but almost linearly in relation to \textit{living space}. This could indicate that in sub-cluster 2, demand for very large properties tapers off, whereas in sub-cluster 5, demand remains stable even for larger properties.

These varying feature-price relationships across sub-clusters highlight the advantages of our two-staged clustering approach with interpretable models at cluster-level. It allows for a more nuanced analysis of cluster-specific trends and leads to a marked improvement in predictive performance. 

\section{Discussion}
\label{sec:discussion}

\subsection{Implications}
This study makes several key contributions to the field of real estate price prediction by addressing the presumed trade-off between model interpretability and predictive performance through a novel two-stage clustering approach. Researchers and practitioners might benefit in various ways. While existing research often applies clustering as a preprocessing step without leveraging its interpretability benefits, our approach integrates clustering directly into the modeling process, enhancing both prediction performance and explainability.

First, we introduce a hierarchical clustering framework that groups properties based on location-specific features before refining clusters using property-specific attributes. This segmentation improves market differentiation and enhances predictive performance.

Second, we evaluate LR and EBM across these clusters, demonstrating that two-stage clustering improves predictive performance by up to $58\%$ for LR and $36\%$ for EBM compared to non-clustered approaches.

Third, we provide cluster-specific insights into price drivers, revealing variations in the influence of factors such as living space and construction year. This enhances model transparency and supports better decision-making for real estate professionals by providing actionable insights.

Finally, our work contributes to the discussion on interpretable machine learning, bridging the gap between high-performance but black-box models and transparent but less predictive approaches \citep{kruschel2025challenging}. Our findings highlight the potential of segmentation-driven modeling for more accurate and actionable property valuations.

\subsection{Limitations \& Future Research}

While our proposal of combining clustering techniques and predictive models for house prices yielded promising results, several inherent limitations affect the predictive quality and reliability.

A prevailing challenge is forecasting prices for luxury properties, which exhibit high heterogeneity and unique features often absent from structured datasets. Features such as aesthetic design elements, special features, or high-end renovations are not adequately captured in tabular data, being only discernible in textual descriptions or images. Additionally, emotional factors and personal buyer preferences play an important role in real estate \citep{ben2014real}, further complicating price predictions. Future research could explore multi-modal inputs incorporating text and image data, and engage with real estate professionals to better understand feature importance and validate the practical utility of model insights.

Moreover, we focused on interpretable models rather than black-box regressors such as RF or XGBoost, aiming to derive inherently interpretable explanations without relying on post-hoc explainers \citep{rudin2019stop}. Future work could investigate model-agnostic interpretability techniques like SHAP \citep{lundberg_local_2020} or LIME \citep{ribeiro2016should} when using black-box models, and conduct comprehensive comparisons with other state-of-the-art methods including advanced ensemble methods and deep learning approaches to better position our methodology within the broader predictive modeling landscape.

Another limitation is the handling of outliers and our location-driven clustering approach. Some properties received exceptionally high valuations despite features suggesting lower prices, indicating that critical attributes may not have been fully captured. Additionally, our clustering was primarily driven by location as the initial splitting criterion, and future research could investigate whether alternative features might be more suitable as the primary clustering dimension.

In addition, our dataset is based on online house property listings, which introduces potential biases as it remains unclear whether listed prices reflect actual transaction prices or include overpriced properties that remained unsold. Access to finalized transaction data would enhance prediction reliability.

Beyond real estate prediction, developing systematic methodologies for determining optimal clustering characteristics and model parameters across different domains represents an important avenue for future research.
\section{Conclusion}
\label{sec:conclusion}

In this study, we applied a two-stage clustering approach to house property listings across Germany for the year 2023. The first clustering stage grouped properties exclusively by location, opting for $k$-means. In the second stage, the clusters were further divided based on additional property features using $k$-means and regression trees.

Our findings demonstrate that clustering improves house price prediction while providing deeper insights into market segments across different locations and property types. The interpretability of clustered models revealed heterogeneous pricing mechanisms, showing that feature impacts on property prices vary across market segments. This segmentation enhances market transparency and benefits buyers, sellers, and real estate professionals by enabling more targeted valuation approaches.

A two-stage $k$-means clustering approach combined with EBM achieved the best predictive performance, improving MAE by up to $36\%$ over non-clustered models. When paired with linear regression, the clustering methodology yielded even greater improvements of approximately $58\%$.

Despite clustering improvements, limitations remain, particularly reduced predictive performance for higher-priced properties due to the greater complexity and heterogeneity of luxury real estate. Future studies could integrate image-based features or textual descriptions to enhance predictions for these high-value properties.

Overall, our study demonstrates that a two-stage clustering approach significantly enhances real estate price prediction performance. By grouping properties into homogeneous categories, this methodology enables targeted feature analysis and yields more precise, transparent valuations. This approach provides valuable insights for real estate professionals, investors, and policymakers through a data-driven pricing framework. Our findings confirm geographical location as the most critical price determinant, while interpretable models ensure transparency and practical applicability, offering clear advantages over black-box approaches where explainability is essential.

\bibliographystyle{agsm}
\bibliography{literature.bib}

\end{document}